\documentclass{article}
\PassOptionsToPackage{numbers, compress}{natbib}

\usepackage[preprint]{neurips_2024}

\usepackage[utf8]{inputenc} %
\usepackage[T1]{fontenc}    %
\usepackage{hyperref}       %
\usepackage{url}            %
\usepackage{booktabs}       %
\usepackage{amsfonts}       %
\usepackage{nicefrac}       %
\usepackage{microtype}      %
\usepackage{xcolor}         %
\usepackage{amsmath}
\usepackage{multirow}
\usepackage{graphicx}
\usepackage{csquotes}
\usepackage{subcaption}
\usepackage[normalem]{ulem}
\usepackage{xspace}

\usepackage{tcolorbox}
\usepackage{colortbl}

\definecolor{eccvblue}{rgb}{0.12,0.49,0.85}
\hypersetup{
  colorlinks=true,
  linkcolor=black,     
  filecolor=black,  
  urlcolor=magenta,      
  citecolor=eccvblue   
}

\useunder{\uline}{\ul}{}
\definecolor{oviscolor}{HTML}{faf2f1}

\usepackage{amssymb}
\usepackage{amsmath,amsfonts}
\usepackage{amsopn}
\usepackage{bm} %
\usepackage{multirow}
\usepackage{flushend}
\usepackage{tabularx}

\newcommand{\vct}[1]{\boldsymbol{#1}} %
\newcommand{\mat}[1]{\boldsymbol{#1}} %

\newcommand{\ProbOpr}[1]{\mathbb{#1}}

\newcommand{\expect}[2]{%
\ifthenelse{\equal{#2}{}}{\ProbOpr{E}_{#1}}
{\ifthenelse{\equal{#1}{}}{\ProbOpr{E}\left[#2\right]}{\ProbOpr{E}_{#1}\left[#2\right]}}} %

\newcommand{\e}{{\vct{e}}}

\newcommand{\vt}{\vct{t}}
\newcommand{\vv}{\vct{v}}
\newcommand{\rr}{{\vct{r}}}

\newcommand{\mW}{\mat{W}}

\newcommand{\eat}[1]{}

\newcommand{\name}{{Ovis}\xspace}

\title{Ovis: Structural Embedding Alignment for\\ Multimodal Large Language Model}

\author{Shiyin Lu$^{1}$  \: Yang Li$^{1}$ \: Qing-Guo Chen$^{1}$ \: Zhao Xu$^{1}$ \\ 
\textbf{Weihua Luo$^{1}$ \: Kaifu Zhang$^{1}$ \: Han-Jia Ye$^{2,3}$}\thanks{Corresponding author, email: yehj@lamda.nju.edu.cn.}  
\vspace*{1mm} \\
\normalfont $^{1}$AI Business, Alibaba Group \: $^{2}$School of Artificial Intelligence, Nanjing University \\ \normalfont $^{3}$National Key Laboratory for Novel Software Technology, Nanjing University \\
\and \url{https://github.com/AIDC-AI/Ovis}
} 

\begin{document}

\maketitle

\begin{abstract}
Current Multimodal Large Language Models (MLLMs) typically integrate a pre-trained LLM with another pre-trained vision transformer through a connector, such as an MLP, endowing the LLM with visual capabilities.
However, the misalignment between two embedding strategies in MLLMs --- the structural textual embeddings based on an embedding look-up table and the continuous embeddings generated directly by the vision encoder --- makes challenges for a more seamless fusion of visual and textual information.
We propose {\name}, a novel MLLM architecture designed to structurally align visual and textual embeddings. {\name} integrates an additional learnable visual embedding table into the visual encoder's process. To capture rich visual semantics, each image patch indexes the visual embedding table multiple times, resulting in a final visual embedding that is a probabilistic combination of the indexed embeddings. This structural approach mirrors the method used for generating textual embeddings.
Empirical evaluations on various multimodal benchmarks show that {\name} outperforms open-source MLLMs of similar parameter scales and even surpasses the proprietary model Qwen-VL-Plus overall.
These results highlight the potential of {\name}' structured visual representation for advancing MLLM architectural design and promoting more effective multimodal learning.
\end{abstract}

\section{Introduction}
The development of Large Language Models (LLMs) is advancing rapidly~\citep{radford2018improving,radford2019language,brown2020language,ouyang2022training,chatgpt,openai2023gpt4}, illuminating the path toward Artificial General Intelligence (AGI). These sophisticated models excel at understanding and generating text with remarkable proficiency~\cite{llama,llama-2,vicuna,alpaca}. However, to approach the complexity and versatility of human intelligence, LLMs must transcend mere textual comprehension. The ability to interpret and understand visual information becomes a critical feature on this journey toward AGI. Consequently, there has been a surge of interest in developing Multimodal Large Language Models (MLLMs) --- models that meld the power of language comprehension and visual perception~\citep{li2022blip,li2023blip,instructblip,minigpt-4,llava,li2023monkey,li2024mini,bai2023qwen,yang2023dawn}.

\begin{figure}[t]
  \centering
  \includegraphics[width=\linewidth]{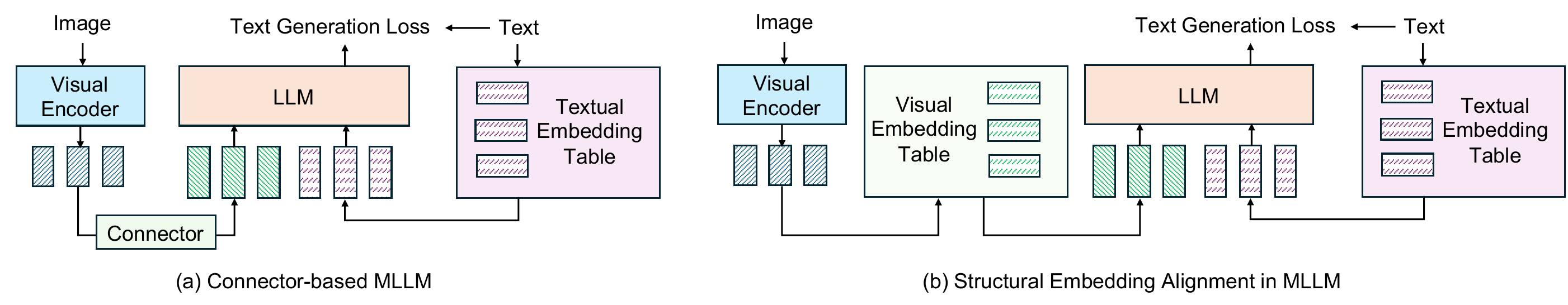}
  \caption{Comparison between different embedding strategies in MLLM.
  In the connector-based approach (a), the connector transforms the visual embeddings into the same dimensional as the textual embedding, where the latter is indexed from a textual embedding table.
  As illustrated in (b), our {\name} leverages an additional visual embedding table to produce structural visual embeddings and align the embedding strategies of two modalities.
  }
  \label{fig:teaser}
  \vspace{-3mm}
\end{figure}

\begin{figure}[t]
  \centering
  
  \begin{subfigure}{0.45\textwidth}
    \includegraphics[width=\linewidth]{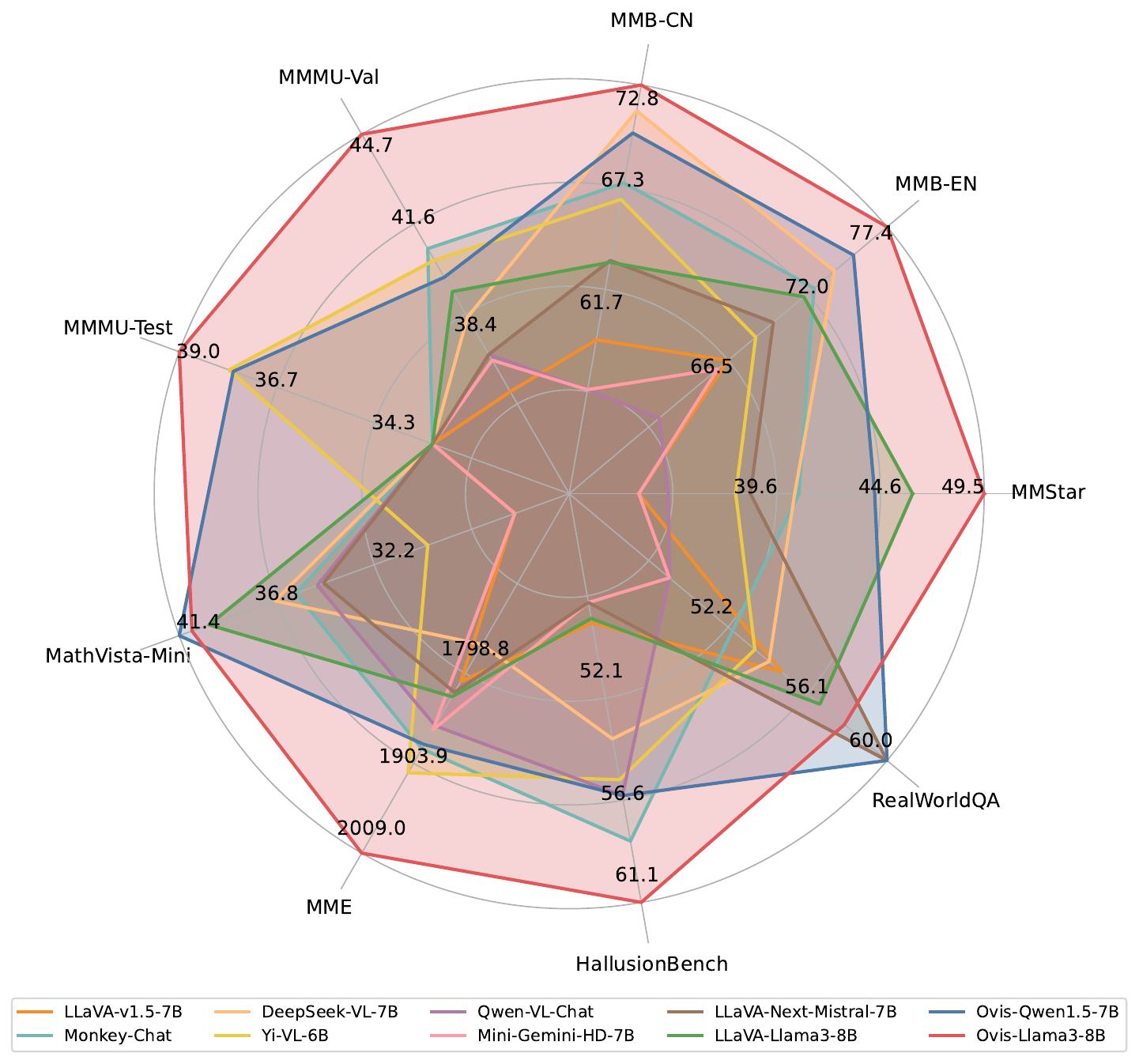}
    \caption{7B tier}
    \label{subfig:radar-chat-a}
  \end{subfigure}
  \hfill
  \begin{subfigure}{0.4215\textwidth}
    \includegraphics[width=\linewidth]{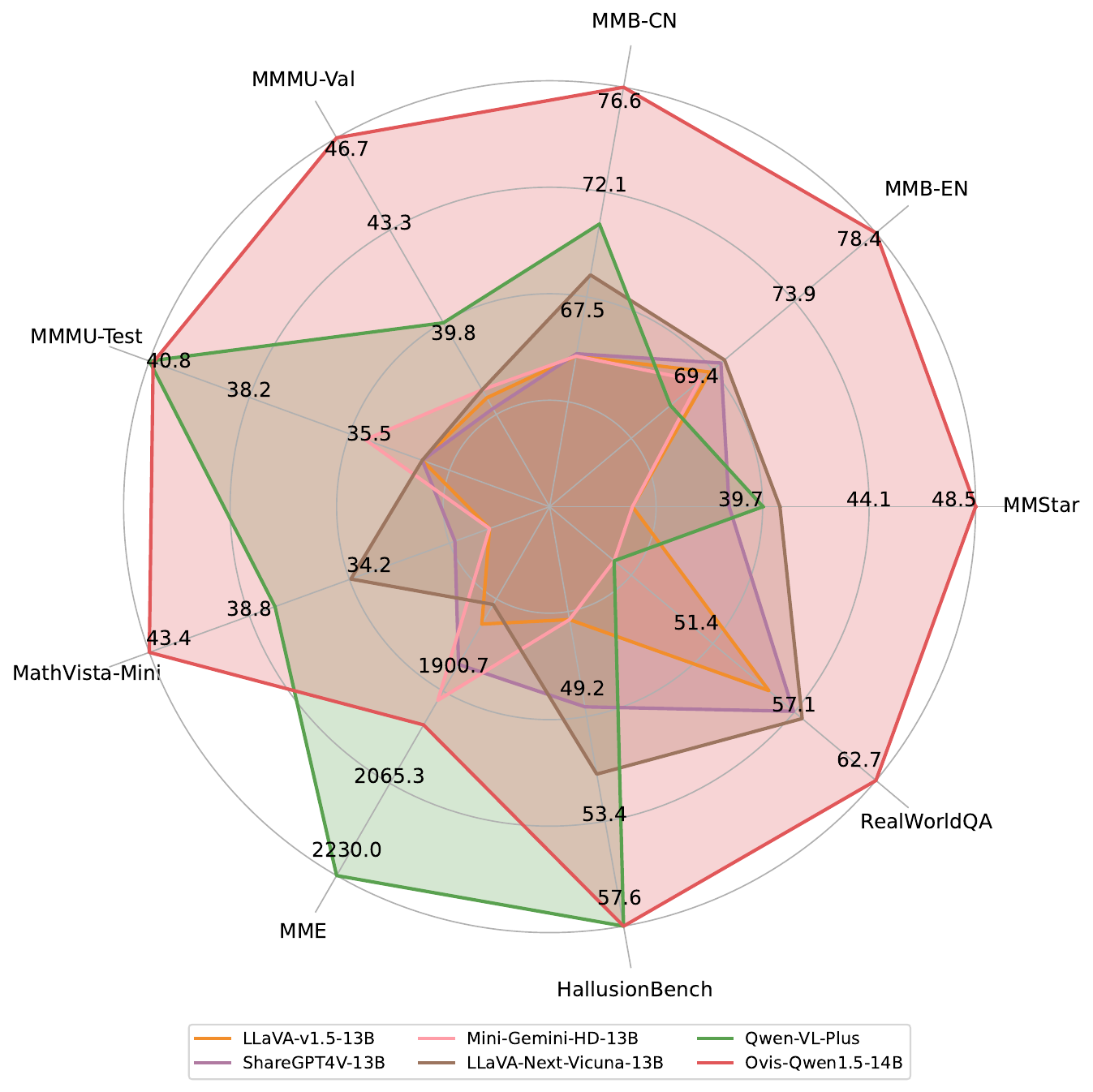}
    \caption{14B tier and Qwen-VL-Plus}
    \label{subfig:radar-chat-b}
  \end{subfigure}

  \caption{{\name} outperforms open-source MLLMs within the same parameter tier in various benchmarks, and {\name}-14B also surpasses the high-resource proprietary model Qwen-VL-Plus overall.}
  \label{fig:radar-chat}
\end{figure}

Instead of directly training the entire MLLMs, current open-source MLLMs primarily derive their visual ability from a pre-trained LLM and a pre-trained vision encoder.
The visual and textual components have different tokenization and embedding strategies. 
Textual embeddings are indexed from the LLM's embedding look-up table, where each ``word'' is mapped to an embedding, via one-hot textual tokens. In contrast, visual embeddings are generated directly by the vision encoder in an unstructured manner. 
To align the dimensions between these two types of embeddings, cross-modal connectors such as MLPs project embeddings into a joint space, allowing all embeddings to serve as inputs to the LLM~\citep{llava,liu2023improved,liu2024llavanext,li2023monkey,li2024mini}. 
Although this architecture only aligns the dimensions of visual and textual embeddings, it has shown promising performance across various vision-language tasks.
Nevertheless, the inherent discrepancy in tokenization and embedding strategies may lead to a potential limitation in the connector-based architecture, so an intuitive question is 
\begin{displayquote}
\vspace{-2mm}
Could we achieve further improvement in MLLMs if we {\em generate visual embeddings in a structured manner} to match the textual embedding strategy in LLMs?
\vspace{-5mm}
\end{displayquote}
We propose a novel MLLM architecture, dubbed ``{\name}'', which assimilates the insights from LLMs to establish structured embeddings of visual input. As illustrated in~\autoref{fig:teaser}, {\name} introduces an additional learnable visual embedding look-up table to transform continuous visual tokens, thus paralleling the structural integrity of its textual counterpart. 
\autoref{fig:radar-chat} demonstrates that {\name} outperforms open-source MLLMs within the same parameter tier across various benchmarks, and {\name}-14B also surpasses the high-resource proprietary model Qwen-VL-Plus overall.

In particular, {\name} incorporates a visual embedding table whose rows correspond to unique visual words, representing distinct visual patterns.
Given the continuous token of a visual patch output by the visual encoder~\citep{dosovitskiy2020image}, {\name} first maps the token into a probabilistic token, revealing its similarity among the entire visual vocabulary set. 
The probabilistic token captures the rich semantics within a single visual patch, which may contain patterns from multiple visual words, effectively treating the visual token as if it were sampled from the visual embedding table based on the distribution.
{\name} subsequently indexes the visual embedding table multiple times based on the probabilistic token, resulting in a final visual embedding that is a combination of the indexed embeddings, i.e., the expectation of embeddings over the whole embedding table.
Therefore, {\name} aligns the visual embedding strategy with the structured nature of its textual counterpart.

The optimization of the visual embedding table and the parameters for generating the probabilistic tokens significantly influence the performance of the MLLM. 
Instead of using an additional autoencoder with vector quantization over images and various other losses, as utilized in previous methods~\citep{van2017neural,esser2021taming,jin2023unified}, {\name} leverages a joint textual generation loss and optimizes the parameters in a three-stage manner. This learning process of {\name} avoids the risk of falling short in vision-language tasks due to the absence of textual guidance.

We implement {\name} with open-source vision transformer and LLM models as backbones and evaluate its performance in diverse multimodal benchmarks. The outcomes demonstrate that {\name} outperforms popular open-source MLLMs within the same parameter tier in the majority of these benchmarks. Specifically, {\name}-8B exhibits a large margin over its competitors, and {\name}-14B consistently surpasses the compared open-source MLLMs. Impressively, {\name}-14B also performs better than the high-resource proprietary model Qwen-VL-Plus \citep{bai2023qwen} overall, and its performance is even on par with the stronger proprietary model Qwen-VL-Max \citep{bai2023qwen} in the general multimodal benchmarks MMStar \citep{chen2024we} and MMBench \citep{mmbench} and several specialized multimodal benchmarks including MathVista \citep{mathvista}, HallusionBench \citep{hallusionbench}, and RealWorldQA \citep{grokv}. These results underscore the superiority and potential of the {\name} architecture. We believe that the demonstrated effectiveness and advantages of {\name} will enhance further investigations into MLLM architectural designs, moving beyond the confines of the connector-based architecture.

\section{Related Work}
\paragraph{Large Language Models.} In recent years, the development of Large Language Models (LLMs) has significantly advanced the field of natural language processing. The debut of GPT-3 \citep{brown2020language} marked a notable surge in performance, especially in few-shot and zero-shot learning scenarios, underscoring the substantial promise of LLMs. This potential was further demonstrated by subsequent enhancements in models such as ChatGPT \citep{chatgpt}, GPT-4 \citep{openai2023gpt4}, Gemini \citep{Gemini, Gemini1.5}, and Claude \citep{Claude}. Concurrently, open-source models have been rapidly evolving, including the LLaMA \citep{llama, llama-2} series, Vicuna \citep{vicuna}, Baichuan \citep{Baichuan-2}, Qwen \citep{qwen-lm}, Mistral \citep{Mistral}, and Yi \citep{Yi}. Notably, the open-source models Llama3 \citep{llama-3} and Mistral-MOE \citep{Mistral-8×22B} have approached and, in some cases, surpassed the performance of closed-source models. Despite these advancements, LLMs inherently lack the capability to process or interpret multimodal data, limiting their application in scenarios requiring an understanding of more than just textual information.

\paragraph{Multimodal Large Language Models.} Multimodal Large Language Models (MLLMs) enhance the capabilities of LLMs by not only understanding and generating text but also interpreting and relating visual elements to textual descriptions \citep{yin2023survey}.
Most open-source MLLMs consist of several components, namely a vision encoder \citep{clip, Eva, Eva2, siglip}, a connector, and an LLM. The type of the connector
can be roughly divided into three categories.
The cross-attention-based methods isolate and integrate visual and text modalities within the LLM, as seen in models like Flamingo \citep{alayrac2022flamingo} and CogVLM \citep{wang2023cogvlm}.
The query-based methods query visual embeddings via a transformer-like architecture and send the obtained visual embeddings along with the text to the LLM, exemplified by Blip-2 \citep{li2023blip}, Instruct-Blip \citep{instructblip}, and Qwen-VL \citep{bai2023qwen}.
The projection-based methods directly project the visual embeddings, align them to the text modality, and uniformly feed the mixed embeddings into the LLM for understanding and generation. This approach is used by models such as LLaVA \citep{llava}, Mini-GPT4 \citep{minigpt-4}, DeepSeek-VL \citep{DeepSeek-VL}, and Mini-Gemini \citep{li2024mini}.
In addition to architecture design, current MLLM research focuses on high-resolution capabilities \citep{HRVDA, InternLM-XComposer2-4KHD, Vary}, miniaturization of MLLMs \citep{MiniCPM, chu2024mobilevlm, lin2024moe}, specialized models (e.g., medical MLLMs \citep{li2023llava}, document MLLMs \citep{OneChart, hu2024mplug}), and the integration of other modalities \citep{wu2023next, X-InstructBLIP, Unified-io-2}.
Our {\name} serves as a new MLLM architecture that departs from the connector-based framework and involves a novel visual tokenizer for structured visual embeddings.

\paragraph{Visual Tokenization.} Tokenizing visual input has been 
explored in various visual tasks~\citep{kingma2013auto}.
VQVAE~\citep{van2017neural} encodes visual input into discrete latent variables, combining the principles of variational autoencoders with vector quantization. This approach facilitates the generation of high-quality and diverse outputs, making it effective for tasks such as image generation and compression. Based on VQVAE, VQGAN~\citep{esser2021taming} incorporates the adversarial training framework of PatchGAN \citep{demir2018patchbased}, enhancing the realism of generated images. Leveraging a visual tokenization strategy similar to VQVAE, BEIT~\citep{bao2021beit, peng2208beit} uses discrete visual tokens during its pre-training phase. In this phase, portions of the input image are masked, and the model predicts the discrete tokens for these masked patches, similar to the masked language modeling in BERT~\citep{Bert}.
Due to the lack of joint modeling with the linguistic modality, there has been scant work combining discretized visual tokens with MLLMs. The discretization of visual tokens has been investigated to link visual output to the input of diffusion models~\citep{Ge2023Planting,Ge2023Making,jin2023unified}, where additional reconstruction loss and decoders are used during training.
A recent method \citep{Multi-modal-Visual-Words} employs a linear head layer to tokenize visual information, which differs from our approach. Specifically, the head layer in \citep{Multi-modal-Visual-Words} is trained solely on vision data in a distilled manner, whereas we optimize the visual head layer using gradients backward from the LLM on vision-language data. Additionally, we propose learning a distinct visual embedding table tailored specifically for visual information, rather than directly using the LLM's textual embedding table to retrieve embeddings for visual tokens as done in \citep{Multi-modal-Visual-Words}.

\section{\name}
\label{sec:method}
In this section, we first review the differences in visual and textual embedding strategies in MLLMs. We then introduce our proposed architecture, {\name}, which incorporates a linear mapping for probabilistic tokens and an additional visual embedding look-up table within LLM.

\begin{figure}[t]
  \centering
  
  \begin{subfigure}{0.45\textwidth}
    \includegraphics[width=\linewidth]{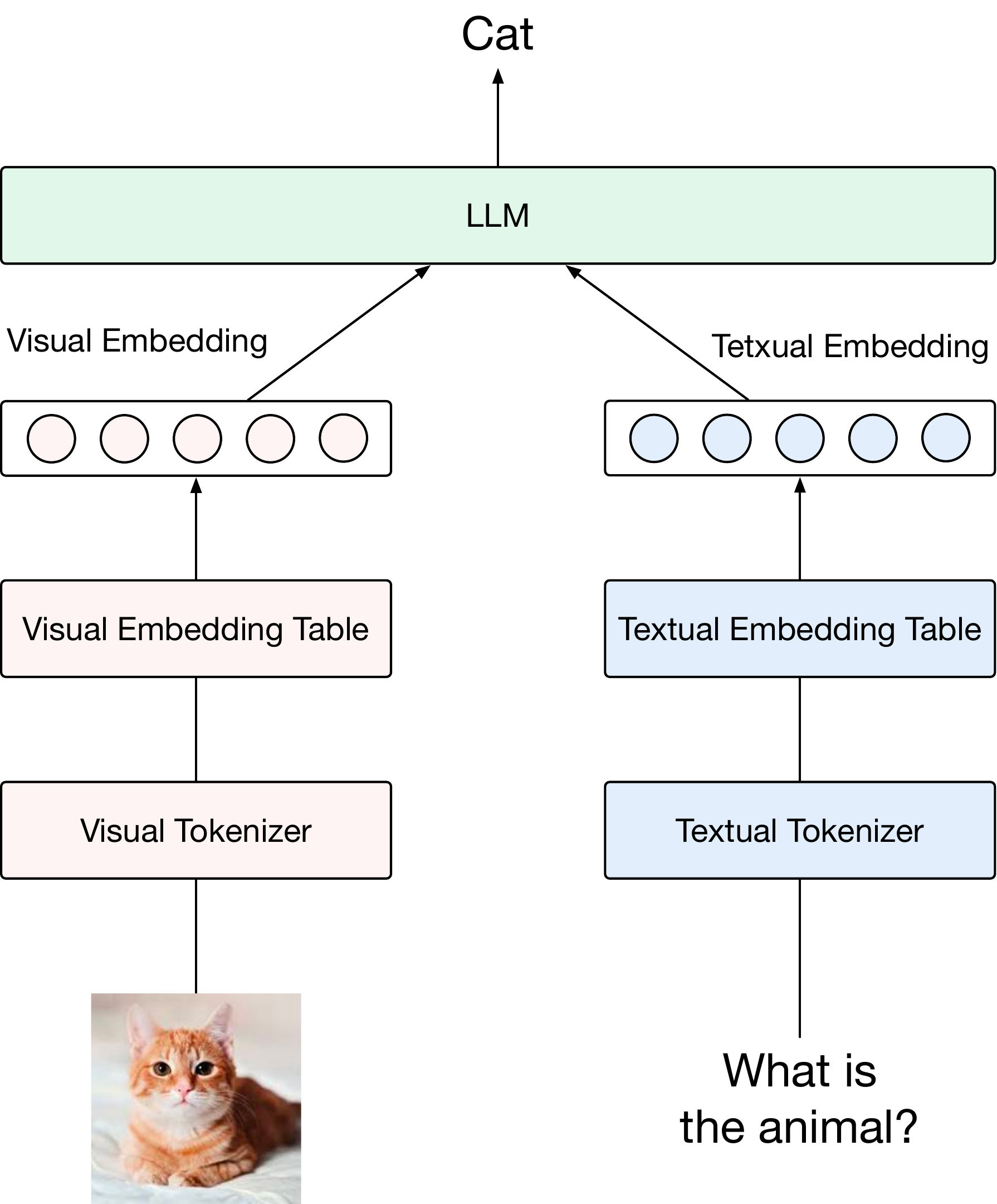}
    \caption{Overall Architecture}
    \label{subfig:genghis-arc-a}
  \end{subfigure}
  \hfill
  \begin{subfigure}{0.45\textwidth}
    \includegraphics[width=\linewidth]{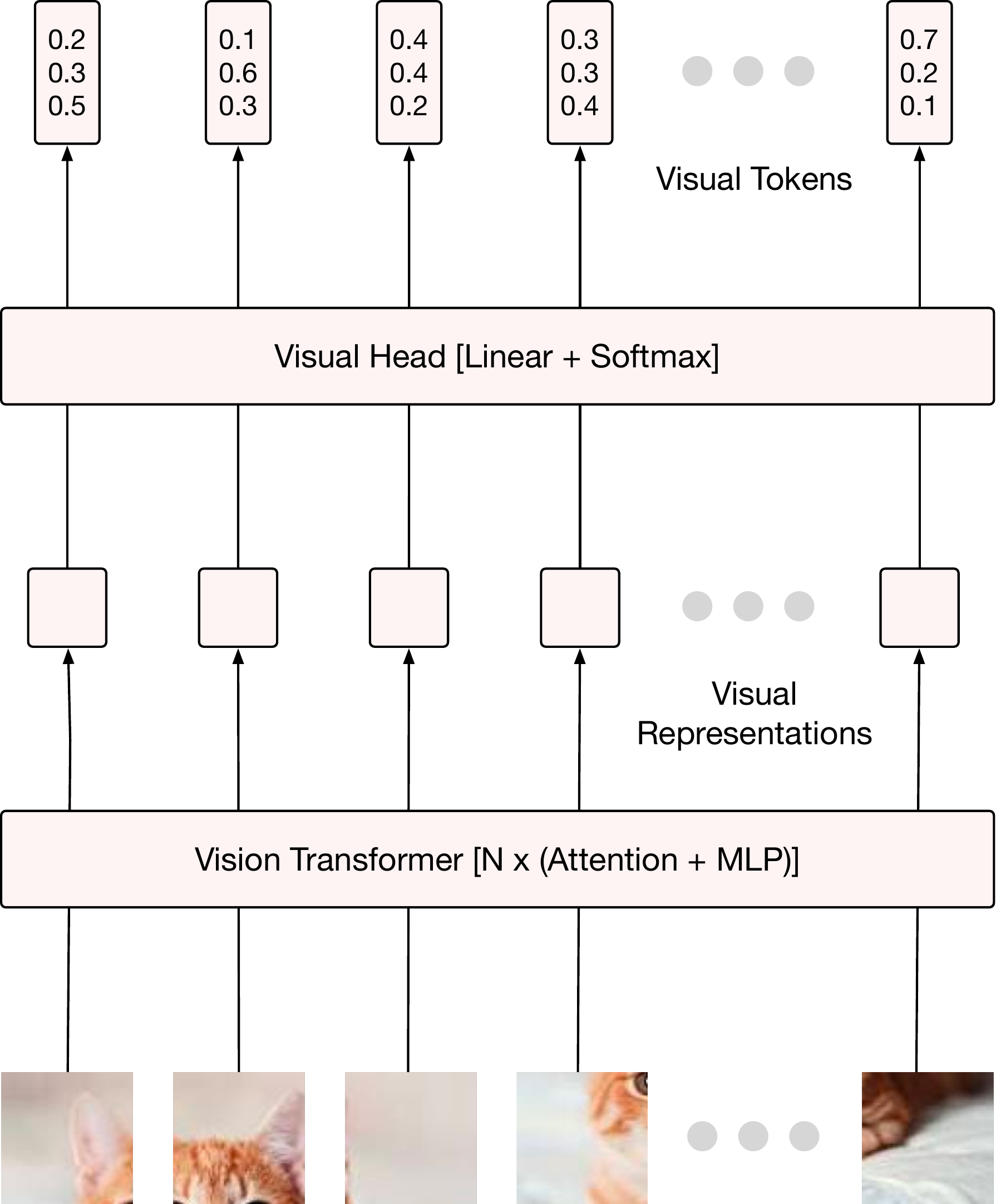}
    \caption{Visual Tokenizer}
    \label{subfig:genghis-arc-b}
  \end{subfigure}
  \begin{subfigure}{\textwidth}
    \includegraphics[width=\linewidth]{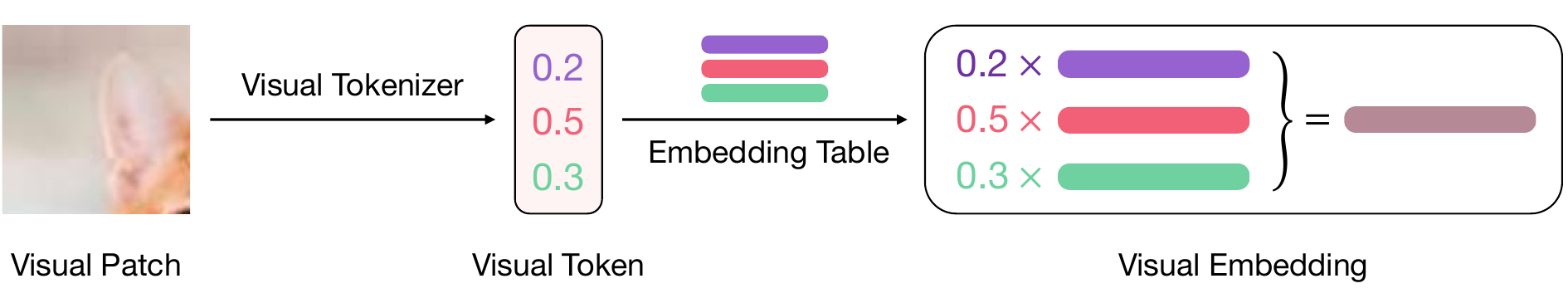}
    \caption{From visual patch to visual embedding}
    \label{subfig:genghis-arc-c}
  \end{subfigure}
  \caption{Illustration of {\name}. Figure (a) shows the whole architecture of {\name}, which contains two embedding tables for visual and textual inputs. Figure (b) illustrates how a visual patch is first mapped to a probabilistic token. Figure (c) demonstrates that the probabilistic token helps select multiple embeddings from the embedding table and output their weighted combination.}
  \label{fig:genghis-arc}
\end{figure}

\subsection{Difference between Visual and Textual Tokens}
Both images and texts are input into the MLLM, and they have diverse tokenization strategies. 

Let $\mathcal{I} \in \mathbb{R}^{C \times W \times H}$ be the pixel value tensor of an image, where $C, W, H$ denote the channel number, width, and height of the image, respectively. 
The image is first divided into a sequence of visual patches $\{\mathcal{P}_i \in \mathbb{R}^{C \times w \times h} \}_{i=1,2,\ldots,n}$, where $w$ and $h$ denote the width and height of the patch, respectively, and $n=\lceil \frac{W}{w} \rceil \lceil \frac{H}{h} \rceil$ is the number of patches.
Given a pre-trained vision transformer (ViT) backbone $g_\theta$ with parameters $\theta$, we then transform the patches into a sequence of visual representations $\{\rr_i \in \mathbb{R}^d \}_{i=1}^n$. 

For the textual input, let $\{\vt_i\}_{i=1}^m$ be the input sequence of textual tokens, which are further processed by an LLM $f_\phi$, parameterized by $\phi$. In MLLM, both visual ($\{\rr_i\}_{i=1}^n$) and textual ($\{\vt_i\}_{i=1}^m$) tokens should be transformed into the same form, and then LLM processes all tokens into an output sequence of textual tokens.
We use $\lambda$ to denote the index of the image indicator token, i.e., $\vt_{\lambda} = \text{\textless image\textgreater}$, and the multimodal input tokens become
\begin{equation}
    [\vt_1, \ldots, \vt_{\lambda-1}, \text{\textless image\textgreater}, \ldots, \vt_m]\;.\label{eq:token_input}
\end{equation}
Since the visual and textual tokens have diverse dimensions, it is difficult to substitute $\text{\textless image\textgreater}$ directly with $\{\rr_i\}_{i=1}^n$. Previous approaches introduce additional linear projection~\cite{llava}, MLP~\cite{liu2023improved}, or transformer~\cite{alayrac2022flamingo,wang2023cogvlm,li2023blip,bai2023qwen} architectures to map visual tokens into the same form as textual ones.

\subsection{Probabilistic Visual Tokens}
Instead of using continuous visual tokens in~\autoref{eq:token_input}, we align the internal tokenization strategies between images and texts to inspire the potential of the MLLM.

To mimic the discrete textual tokens, we use a linear head $\mathbb{R}^{K \times d}$ to transform the concrete visual tokens. Assuming $K$ is the visual vocabulary size, i.e., the number of unique visual words, then given a visual token $\rr_i$, we first transform $\rr_i$ into a $(K-1)$-dimensional probability simplex $\Delta^K$ by a linear projection followed by a softmax normalization:
\begin{equation}
    \vv_i = \text{softmax}(\mW\rr_i)\quad \mW \in \mathbb{R}^{K \times d}.\label{eq:linear_projection}
\end{equation}
We set $\vv_i\in\Delta^K$ as a kind of probabilistic token, which is a probability distribution over the visual vocabulary containing $K$ visual words. If $\rr_i$ is more related to certain patterns, the corresponding elements in $\vv_i$ should be larger.

\noindent{\bf Remark}. Given a visual embedding table, we associate each visual word with its prototype $\{\boldsymbol{w}_i\in\mathbb{R}^d\}_{i=1}^K$. To match a continuous visual token with the $K$ visual words in the embedding table, we leverage the inner product to calculate their similarity value. 
\autoref{eq:linear_projection}
 is the normalized similarity between $\rr_i$ and all visual words.
 
\subsection{Visual Embedding Table}
In LLMs, it is a common practice to employ a textual embedding table, which maps each word in the vocabulary to an embedding vector. 
For each textual token $\vt_i$ in the one-hot form, its embedding $T_i \in \mathbb{R}^{d'}$ is the row of the textual embedding table indicated by the non-zero index in $\vt_i$.

Analogously, we introduce an additional visual embedding table, where each visual word (each row) is associated with an embedding vector $\e_k \in \mathbb{R}^{d'}$ with $d'$ being the embedding dimension.
To make the embeddings of visual and textual tokens have compatible shapes, we simply set the dimension of the visual embedding table to be the same as that of the textual embedding table. 

Accordingly, the embedding of each visual token $\vv_i$ can be derived based on the probabilistic token:
\begin{equation}
    \label{eq:visual-embedding-def}
    V_i = \sum_{k=1}^{K} v_{i,k} \e_k \in \mathbb{R}^{d'}\;,
\end{equation}
where $v_{i,k}$ denotes the $k$-th component of $\vv_i$. 
On the other hand, since $\vv_i \in \Delta^K $, the above formula can be rewritten as $V_i = \mathbb{E}_{k \sim \vv_i} [\e_k]$, which is an expectation of the visual word's embedding, with the visual word drawn from $\vv_i$. 
In other words, we assume that the visual embedding could be sampled from the discrete visual embedding table based on the probabilistic token $\vv_i$ of the patch. 

\noindent{\bf Remark}. Considering the polysemous nature of a visual patch $\rr_i$, assigning it only one visual word from the embedding table indexed by $\arg\max_{j\in\{1,\ldots,K\}} \vv_{ij}$ may neglect the rich semantics within the patch.
To address this, we link the patch with {\em multiple} visual words at a time, as indicated by the non-zero elements in $\vv_i$, which represent the correlation between the patch and $K$ visual words.
We then use the weighted combination $V_i$ of the selected visual words as the final patch embedding.
In other words, multiple visual embeddings are indexed from the embedding table based on the values in $\vv_i$, and the weighted average of these embeddings serves as the final output of the visual embedding module.   
Experiments in~\autoref{sec:appendix-sparsity} validate the sparsity of $\vv_i$. 
The visual embedding $V_i$ captures the rich visual semantics and keeps its generation process similar to its textual counterpart simultaneously.

\subsection{Training Strategy of {\name}}
Both visual and textual embeddings are combined as the input into the LLM. 
In particular, we feed the following multimodal input embedding sequence
\begin{equation}
  [T_1, \ldots, T_{\lambda-1}, V_1, \ldots, V_n, T_{\lambda+1}, \ldots, T_m]  
\end{equation}
to the LLM. All tokens not only have the same dimensionality but are also generated in a similar manner with embedding tables.
LLM will output a textual token sequence $o_1, \ldots, o_l$. 

{\name} is trained in a three-stage fashion and optimized throughout with the textual generation loss, i.e., the cross-entropy between the textual tokens of {\name} output and that of the ground-truth text. The stages differ in their trainable parameters and the types of training data.

\noindent{\bf Stage 1.} 
We freeze all parameters of LLM, as well as most parameters in the visual encoder $g$, an open-source pre-trained ViT backbone. 
We randomly re-initialize the parameters within the last block of $g$, and using visual caption datasets such as COYO \citep{kakaobrain2022coyo-700m} to train the re-initialized parameters as well as the projection $\mW$ and the visual embedding table $\{\e_k\}_{k=1}^K$ of {\name}. 
For each image in the caption dataset, we construct a training sample with input as ``\textless image\textgreater 's caption: '' and label it as ``CAPTION'', where CAPTION denotes the image's caption. 

\noindent{\bf Stage 2.} In this stage, we advance the training of {\name}'s $\mW$, the visual embedding table $\{\e_k\}_{k=1}^K$, and all parameters in the vision encoder $g$.
The LLM is still frozen.
In contrast to the caption samples used in the first stage, we leverage visual description datasets such as ShareGPT4V-Pretrain \citep{chen2023sharegpt4v}, which consist of training samples structured as dialogues that describe images. 

\noindent{\bf Stage 3.} After endowing {\name} with visual perception capabilities through training in Stage 1 and Stage 2, this stage focuses on multimodal instruction learning. The goal is to equip {\name} with the ability to follow multimodal instructions. To this end, we unfreeze the LLM module and train {\name}'s entire set of parameters on multimodal instruction datasets, such as LLaVA-Finetune \citep{liu2023improved}.

\section{Experiments}
\label{sec:exp}
In this section, we provide empirical results to demonstrate the effectiveness of the proposed MLLM architecture {\name}.\footnote{Qualitative results are presented in \autoref{sec:qr}.}
\subsection{Experimental Setup}
\paragraph{Implementation Details.}
{\name} encompasses three configurations: the LLM module, the ViT backbone, and the visual vocabulary size. We incorporate popular open-source LLMs (Qwen1.5-Chat \citep{qwen-lm} and Llama3-Instruct \citep{llama-3}) and ViTs (Clip-ViT-L/14@336px \citep{clip}) into {\name}. The visual vocabulary size is set to $2^{17} = 131,072$, a value comparable to LLMs' textual vocabulary size. To facilitate community use and future innovation, the {\name} architecture and its training code are built upon the widely-used Transformers \citep{wolf-etal-2020-transformers} and DeepSpeed \citep{deepspeed} packages. We detail the training hyper-parameters for each stage in \autoref{tab:hp} of \autoref{sec:td}.

\paragraph{Training Datasets.} {\name} is trained predominantly on open-source datasets, supplemented by a smaller proportion of in-house datasets. The datasets employed can be categorized into three groups: visual captions, visual descriptions, and multimodal instructions, which are respectively utilized in Stage 1, 2, and 3 of the training process. The visual caption dataset is extracted from the COYO dataset based on the similarity between the image and its caption. We leveraged the COYO dataset's provided ``clip-similarity-vitb32'' and ``clip-similarity-vitl14'' scores for this purpose. Specifically, we select all entries from the COYO dataset with both similarity metrics exceeding $0.36$. The visual description datasets and multimodal instruction datasets are all converted into the same format as LLaVA-Finetune \citep{liu2023improved}. Our in-house datasets are available at \url{https://huggingface.co/datasets/AIDC-AI/Ovis-dataset}. We describe the constructions and present several samples from the in-house datasets in \autoref{sec:id1} and \autoref{sec:id2}. Statistics of the training dataset are reported in \autoref{tab:data-stats} of \autoref{sec:td}.

\subsection{Main Results}
\begin{table}[t]
\centering
\caption{Comparison with popular open-source and proprietary MLLMs on general multimodal benchmarks. MMBench is shortened to MMB due to width limitations. MMMU-V and MMMU-T denote the validation and test splits, respectively. The GPT4V-HR denotes GPT4V in the high-resolution mode. Within each parameter tier, the top-performing model is highlighted in bold, while the runner-up is marked with an underscore.}
\vspace{\baselineskip}
\label{tab:general-bench}
\begin{tabular}{@{}lccccc@{}}
\toprule
\multicolumn{1}{l|}{MLLM}                  & MMStar & MMB-EN & MMB-CN & MMMU-V & MMMU-T \\ \midrule
\multicolumn{6}{c}{\textit{open-source models in 7B tier}}                                        \\ \midrule
\multicolumn{1}{l|}{InstructBLIP-7B \citep{instructblip}}       & 32.7   & 33.9    & 23.9       & 30.6     & 33.8      \\
\multicolumn{1}{l|}{VW-LMM-PIF-7B \citep{Multi-modal-Visual-Words}}       & -   & 65.2   & 53.5       & -     & -      \\
\multicolumn{1}{l|}{LLaVA-v1.5-7B \citep{liu2023improved}}         & 33.1   & 66.5    & 59.0       & 35.7     & -         \\
\multicolumn{1}{l|}{ShareGPT4V-7B \citep{chen2023sharegpt4v}}         & 35.7   & 67.6    & 60.7       & 37.2     & -         \\
\multicolumn{1}{l|}{Monkey \citep{li2023monkey}}                & 37.0   & 59.6    & 54.7       & 38.9     & -         \\
\multicolumn{1}{l|}{Monkey-Chat \citep{li2023monkey}}           & 40.7   & 72.4    & 67.5       & {\ul 40.7}     & -         \\
\multicolumn{1}{l|}{DeepSeek-VL-7B \citep{DeepSeek-VL}}        & 40.5   & 73.8    & {\ul 71.4}       & 38.3     & -         \\
\multicolumn{1}{l|}{Yi-VL-6B \citep{Yi}}              & 37.7   & 68.4    & 66.6       & 40.3     & {\ul 37.8}      \\
\multicolumn{1}{l|}{Qwen-VL-Chat \citep{bai2023qwen}}          & 34.5   & 61.8    & 56.3       & 37.0     & 32.9      \\
\multicolumn{1}{l|}{Mini-Gemini-7B \citep{li2024mini}}        & -      & 69.3    & -          & 36.1     & 32.8      \\
\multicolumn{1}{l|}{Mini-Gemini-HD-7B \citep{li2024mini}}     & -      & 65.8    & -          & 36.8     & 32.9      \\
\multicolumn{1}{l|}{LLaVA-Next-Vicuna-7B \citep{liu2024llavanext}}  & 37.6   & 69.2    & 62.3       & 37.6     & -         \\
\multicolumn{1}{l|}{LLaVA-Next-Mistral-7B \citep{liu2024llavanext}} & 38.4   & 69.6    & 63.3       & 37.0     & -         \\
\multicolumn{1}{l|}{LLaVA-Llama3-8B \citep{2023xtuner}} & {\ul 46.1}   & 71.7    & 63.2       & 39.2     & -         \\
\multicolumn{1}{l|}{VILA1.5-Llama3-8B \citep{lin2023vila}} & - & 72.3 & 66.2 & 36.9 & 36.0 \\
\rowcolor{oviscolor} \multicolumn{1}{l|}{Ovis-Qwen1.5-7B}          & 44.3   & {\ul 75.1}    & 70.2       & 39.7     & 37.7      \\
\rowcolor{oviscolor} \multicolumn{1}{l|}{Ovis-Llama3-8B}          & \textbf{49.5}   & \textbf{77.4}    & \textbf{72.8}       & \textbf{44.7}     & \textbf{39.0}      \\ \midrule
\multicolumn{6}{c}{\textit{open-source models in 14B tier}}                                       \\ \midrule
\multicolumn{1}{l|}{PandaGPT-13B \citep{pandagpt}}          & 25.6   & 42.5    & 32.0       & 32.9     & -         \\
\multicolumn{1}{l|}{LLaVA-v1.5-13B \citep{liu2023improved}}        & 34.3   & 69.2    & 65.0       & 37.0     & 33.6      \\
\multicolumn{1}{l|}{ShareGPT4V-13B \citep{chen2023sharegpt4v}}        & 38.3   & 69.8    & 65.1       & 36.6     & -         \\
\multicolumn{1}{l|}{Mini-Gemini-13B \citep{li2024mini}}       & -      & 68.5    & -          & {\ul 38.1}     & 33.5      \\
\multicolumn{1}{l|}{Mini-Gemini-HD-13B \citep{li2024mini}}    & -      & 68.6    & -          & 37.3     & {\ul 35.1}      \\
\multicolumn{1}{l|}{LLaVA-Next-Vicuna-13B \citep{liu2024llavanext}} & {\ul 40.4}   &  70.0    & {\ul 68.5}       & 37.3     & -         \\
\multicolumn{1}{l|}{VILA1.5-13B \citep{lin2023vila}} & - & {\ul 74.9} & 66.3 & 37.9 & 33.6 \\
\rowcolor{oviscolor}
\multicolumn{1}{l|}{Ovis-Qwen1.5-14B}         & \textbf{48.5}   & \textbf{78.4}    & \textbf{76.6}       & \textbf {46.7}     & \textbf{40.7}      \\ \midrule
\multicolumn{6}{c}{\textit{proprietary models}}                                                   \\ \midrule
\multicolumn{1}{l|}{GPT4V \citep{openai2023gpt4}}              & 52.9   & 80.8    & 79.1       & 62.3     & 55.7      \\
\multicolumn{1}{l|}{GPT4V-HR \citep{openai2023gpt4}}              & 56.0   & 81.0    & 80.2       & 61.7     & -         \\
\multicolumn{1}{l|}{Gemini-Pro \citep{Gemini}}             & 38.6   & 73.6    & 74.3       & 49.0     & -         \\
\multicolumn{1}{l|}{Qwen-VL-Plus \citep{bai2023qwen}}          & 39.7   & 67.0    & 70.7       & 39.8     & 40.8      \\
\multicolumn{1}{l|}{Qwen-VL-Max \citep{bai2023qwen}}           & 49.5   & 77.6    & 75.7       & 52.0     & 46.8      \\ \bottomrule
\end{tabular}
\end{table}

\begin{table}[t]
\centering
\caption{Comparison with popular open-source and proprietary MLLMs on specialized multimodal benchmarks. The GPT4V-HR denotes GPT4V in the high-resolution mode. We report the sum of perception and cognition scores for MME and the QuestionAcc score for HallusionBench. Within each parameter tier, the top-performing and runner-up models are in bold and underscored, respectively.}
\vspace{\baselineskip}
\label{tab:specialized-bench}
\begin{tabular}{@{}lcccc@{}}
\toprule
MLLM                                       & MathVista-Mini & MME  & HallusionBench & RealWorldQA \\ \midrule
\multicolumn{5}{c}{\textit{open-source models in 7B tier}}                                        \\ \midrule
\multicolumn{1}{l|}{InstructBLIP-7B \citep{instructblip}}       & 24.4           & 1391 & 53.6           & 36.9        \\
\multicolumn{1}{l|}{LLaVA-v1.5-7B \citep{liu2023improved}}         & 25.6           & 1808 & 48.8           & 54.8        \\
\multicolumn{1}{l|}{ShareGPT4V-7B \citep{chen2023sharegpt4v}}         & 26.5           & {\ul 1915} & 48.8           & 54.9        \\
\multicolumn{1}{l|}{Monkey \citep{li2023monkey}}                & 33.5           & 1760 & 55.1           & 51.6        \\
\multicolumn{1}{l|}{Monkey-Chat \citep{li2023monkey}}           & 35.9           & 1887 & {\ul 58.4}           & 52.4        \\
\multicolumn{1}{l|}{DeepSeek-VL-7B \citep{DeepSeek-VL}}        & 36.9           & 1765 & 53.9           & 54.2        \\
\multicolumn{1}{l|}{Yi-VL-6B \citep{Yi}}              & 29.7           & {\ul 1915} & 55.7           & 53.5        \\
\multicolumn{1}{l|}{Qwen-VL-Chat \citep{bai2023qwen}}          & 34.9           & 1860 & 56.4           & 49.3        \\
\multicolumn{1}{l|}{Mini-Gemini-7B \citep{li2024mini}}        & -              & 1839 & -              & -           \\
\multicolumn{1}{l|}{Mini-Gemini-HD-7B \citep{li2024mini}}     & -              & 1865 & -              & -           \\
\multicolumn{1}{l|}{LLaVA-Next-Vicuna-7B \citep{liu2024llavanext}}  & 31.5           & 1769 & 47.2           & 57.8        \\
\multicolumn{1}{l|}{LLaVA-Next-Mistral-7B \citep{liu2024llavanext}} & 34.6           & 1821 & 47.9           & \textbf{60.0}        \\
\multicolumn{1}{l|}{LLaVA-Llama3-8B \citep{2023xtuner}} & 40.0           & 1826 & 48.6           & 56.7        \\
\rowcolor{oviscolor} \multicolumn{1}{l|}{Ovis-Qwen1.5-7B}          & \textbf{41.4}           & 1882 & 56.4           & \textbf{60.0}        \\
\rowcolor{oviscolor} \multicolumn{1}{l|}{Ovis-Llama3-8B}          & {\ul 40.8}           & \textbf{2009} & \textbf{61.1}           & {\ul 57.9}        \\ \midrule
\multicolumn{5}{c}{\textit{open-source models in 14B tier}}                                       \\ \midrule
\multicolumn{1}{l|}{PandaGPT-13B \citep{pandagpt}}          & 25.0           & 1072 & 43.1           & 32.8        \\
\multicolumn{1}{l|}{LLaVA-v1.5-13B \citep{liu2023improved}}        & 27.7           & 1781 & 45.3           & 55.3        \\
\multicolumn{1}{l|}{ShareGPT4V-13B \citep{chen2023sharegpt4v}}        & 29.3           & 1853 & 48.8           & 57.0        \\
\multicolumn{1}{l|}{Mini-Gemini-13B \citep{li2024mini}}       & -              & 1887 & -              & -           \\
\multicolumn{1}{l|}{Mini-Gemini-HD-13B \citep{li2024mini}}    & -              & {\ul 1917} & -              & -           \\
\multicolumn{1}{l|}{LLaVA-Next-Vicuna-13B \citep{liu2024llavanext}} & {\ul 34.1}           & 1746 & {\ul 51.5}           & {\ul 57.6}        \\
\rowcolor{oviscolor} \multicolumn{1}{l|}{Ovis-Qwen1.5-14B}         & \textbf{43.4}           & \textbf{1961} & \textbf{57.6}           & \textbf{62.7}        \\ \midrule
\multicolumn{5}{c}{\textit{proprietary models}}                                                   \\ \midrule
\multicolumn{1}{l|}{GPT4V \citep{openai2023gpt4}}                 & 51.4           & 2038 & 60.1           & 61.4        \\
\multicolumn{1}{l|}{GPT4V-HR \citep{openai2023gpt4}}              & 54.7           & 2070 & 62.1           & 68.0        \\
\multicolumn{1}{l|}{Gemini-Pro \citep{Gemini}}             & 46.5           & 2149 & 63.7           & 60.4        \\
\multicolumn{1}{l|}{Qwen-VL-Plus \citep{bai2023qwen}}          & 37.6           & 2230 & 57.6           & 44.6        \\
\multicolumn{1}{l|}{Qwen-VL-Max \citep{bai2023qwen}}           & 43.4           & 2282 & 57.7           & 61.3        \\ \bottomrule
\end{tabular}
\end{table}

We evaluate {\name} across a variety of benchmarks, covering both general multimodal capabilities benchmarks (MMMU \citep{yue2023mmmu}, MMBench-EN \citep{mmbench}, MMBench-CN \citep{mmbench}, and MMStar \citep{chen2024we}), as well as benchmarks for more specialized multimodal tasks (MathVista-Mini \citep{lu2024mathvista}, MME \citep{mme}, HallusionBench \citep{hallusionbench}, and RealWorldQA \citep{grokv}). The evaluation is performed using the VLMEvalKit package \citep{2023opencompass}. The comparison between {\name}' benchmark performance with that of popular open-source MLLMs and leading proprietary models is summarized in \autoref{tab:general-bench} and \autoref{tab:specialized-bench}, where the benchmark scores of the compared models are mainly derived from VLMEvalKit for consistency. MLLMs with no specialized multimodal benchmark performance reported are not included in \autoref{tab:specialized-bench} for conciseness.

It can be seen that Ovis-8B outperforms the open-source models of similar size across the majority of benchmarks. Ovis-14B not only excels in all benchmarks but also surpasses the high-resource proprietary model Qwen-VL-Plus in most benchmarks. In the vision-indispensable multi-modal benchmark MMStar, Ovis-8B exhibits a large margin over the compared open-source MLLMs, highlighting its advantage in utilizing visual information. Ovis also achieves leading results in the highly challenging college-level MMMU benchmark, demonstrating strong visual comprehension and reasoning abilities. The MMBench-EN and MMBench-CN benchmarks differ only in the language \citep{mmbench}. While Ovis' training dataset contains very few non-English samples, Ovis performs well in both versions. Ovis-14B achieves consistently outstanding performance in MMBench-EN and MMBench-CN, suggesting that Ovis' advantage in multimodal capabilities is not limited to English but can extend to another language like Chinese as well.

Turning attention to the specialized multimodal benchmarks, we find that Ovis enjoys better multimodal capabilities in math and logical reasoning than open-source competitors, as demonstrated by its notable performance in the MathVista-Mini benchmark. While Ovis only employs a 336px ViT backbone and is not equipped with high-resolution-boosted techniques such as the dynamic high resolution used in LLaVA-Next \citep{liu2024llavanext} and the dual vision encoders used in Mini-Gemini-HD \citep{li2024mini}, Ovis exhibits impressive performance in the RealWorldQA benchmark, which is comprised of real-world visual tasks with high-resolution images (e.g., 1080P). Notably, Ovis-14B’s RealWorldQA score is even higher than the leading proprietary model GPT4V, illustrating its outstanding multimodal capabilities in solving practical visual tasks. In the MME and hallucination benchmarks, Ovis-8B and Ovis-14B perform the best within the 7B and 14B tier, respectively. This implies that Ovis' strong visual understanding and reasoning abilities are accompanied by a lower rate of hallucination, a highly desirable trait for the deployment of MLLMs in critical scenarios such as medicine.

\subsection{Ablation Study}
\begin{table}[t]
\centering
\caption{Comparison between Ovis and the conventional connector-based architecture, both employing Qwen1.5-7B-Chat and Clip-ViT-L/14@336px as backbones and trained on the same datasets. Due to width limitations, MMBench-EN and MMBench-CN are merged into a single column, as are MMMU-V and MMMU-T. MathVista-Mini, HallusionBench, and RealWorldQA are shortened to Math, HB, and RWQA, respectively.}
\vspace{\baselineskip}
\label{tab:arch-ablation}
\begin{tabular}{@{}lccccccccc@{}}
\toprule
Architecture        & MMStar & MMBench & MMMU & Math & MME   & HB & RWQA \\ \midrule
Connector & 41.1   & 71.0    / 65.2        & 34.8 / 33.8      & 36.3      & 1757  & 54.0           & 56.1        \\
Ovis & 44.3   & 75.1    / 70.2        & 39.7     / 37.7      & 41.4      & 1882  & 56.4           & 60.0        \\ \midrule
Improvement& 7.8\%  & 5.8\%   / 7.7\%       & 14.1\%   / 11.5\%    & 14.0\%    & 7.1\% & 4.4\%          & 7.0\%      \\ \bottomrule
\end{tabular}
\end{table}

To further elucidate the advantages of Ovis' architectural design, we conduct a comparative experiment between Ovis-7B and a connector-based MLLM utilizing identical LLM and ViT backbones as Ovis-7B. Following \citep{liu2023improved}, we implement the connector as a two-layer MLP with GELU activation. The hidden size of the MLP is configured to match Ovis-7B's visual vocabulary size, ensuring parity in parameter count between the connector-based MLLM and Ovis-7B. We train the connector-based MLLM on the same datasets as Ovis-7B, adhering to the training paradigm outlined in \citep{liu2023improved}. The experimental results are summarized in \autoref{tab:arch-ablation}. Remarkably, Ovis consistently outperforms the connector-based architecture across all benchmark evaluations, achieving an impressive 8.8\% performance margin on average. Given the identical parameter counts, backbones, and training datasets, the results compellingly advocate for the efficacy of Ovis' architectural design.

\section{Conclusion}
We emphasize the necessity of structurally aligning visual embeddings with the textual counterparts, considering their different tokenization and embedding strategies in MLLMs. In {\name}, we introduce an additional visual embedding look-up table. 
Image patches are mapped into probabilistic tokens, which then index the visual embedding table and are transformed into a structural manner similar to textual embeddings.
Empirical evaluations across various multimodal benchmarks validate Ovis' effectiveness, demonstrating that it outperforms open-source MLLMs of similar parameter scales as well as the proprietary model Qwen-VL-Plus.

\section{Broader Impact and Limitations}
\label{sec:bi-li}
\paragraph{Broader Impact.} As a powerful multimodal large language model architecture, {\name} has the potential to benefit a wide range of users through enhanced interactions between visual content and textual analysis. 
However, it is crucial to acknowledge the potential negative impacts associated with {\name}, such as the risk of hallucination, wherein {\name} may generate misleading or incorrect information, potentially leading to misinformation. Furthermore, {\name} also suffers from biases and potential harms, a common issue among generative models. These potential adverse effects could be mitigated through content moderation mechanisms and transparent model developments.

\paragraph{Limitations.} While {\name} has demonstrated promising performance, its efficacy in handling visual tasks with high-resolution images is limited due to the absence of high-resolution-boosted techniques. Moreover, {\name} is trained solely with single-image samples, posing challenges when confronted with scenarios requiring visual understanding across multiple images.
Considerable research efforts have been dedicated to these areas \cite{liu2024llavanext, li2024mini,lin2023vila}, primarily within the connector-based framework. Drawing inspiration from these researches, we plan to enhance {\name}' capacity to better handle high-resolution images and process multi-image inputs in future iterations.

\bibliographystyle{plainnat}
\bibliography{ref}

\appendix
\clearpage{}
\section{Qualitative Results}
\label{sec:qr}
\begin{figure}[h]
  \centering
\begin{subfigure}{\textwidth}
    \includegraphics[width=\linewidth]{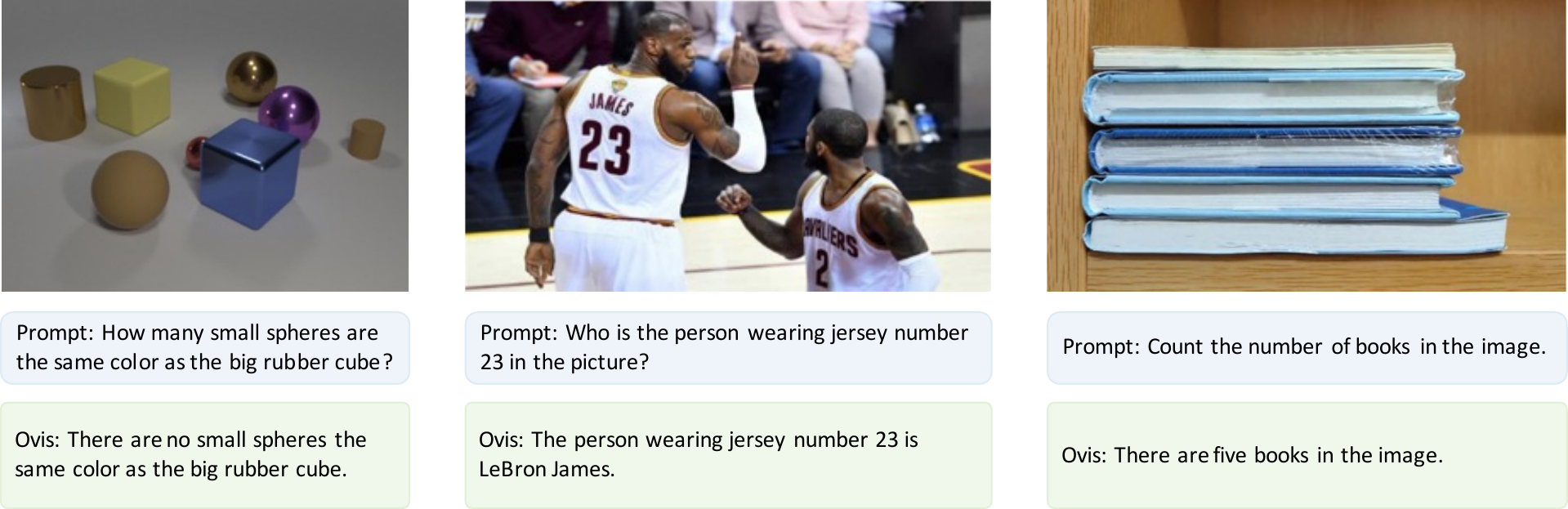}
    \caption{Perception \citep{li2024mini,li2023monkey,fu2023gemini}}
    \label{fig:ovis-sample-perception}
  \end{subfigure}

  \hspace{1cm}
  
  \begin{subfigure}{\textwidth}
    \includegraphics[width=\linewidth]{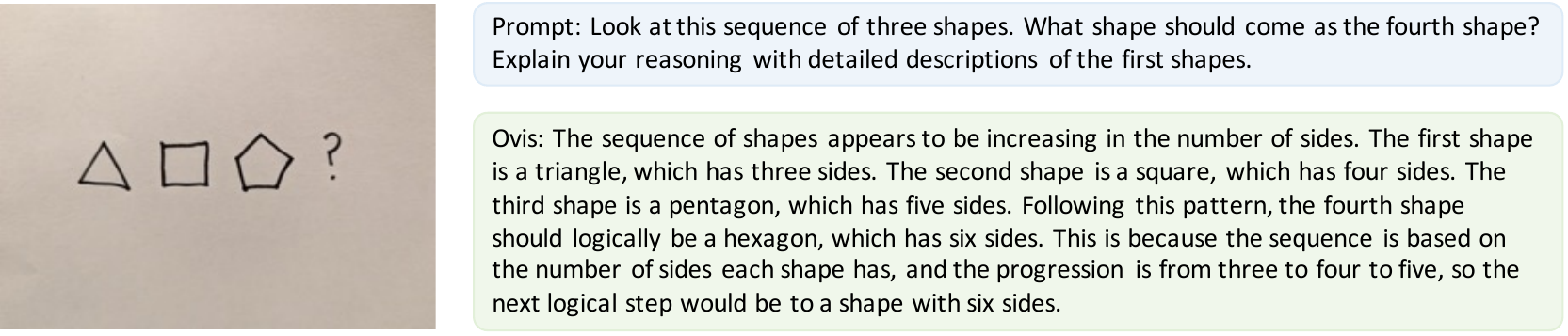}
    \caption{Reasoning \citep{Gemini}}
    \label{fig:ovis-samples-b}
  \end{subfigure}

  \hspace{1cm}
  
  \begin{subfigure}{\textwidth}
    \includegraphics[width=\linewidth]{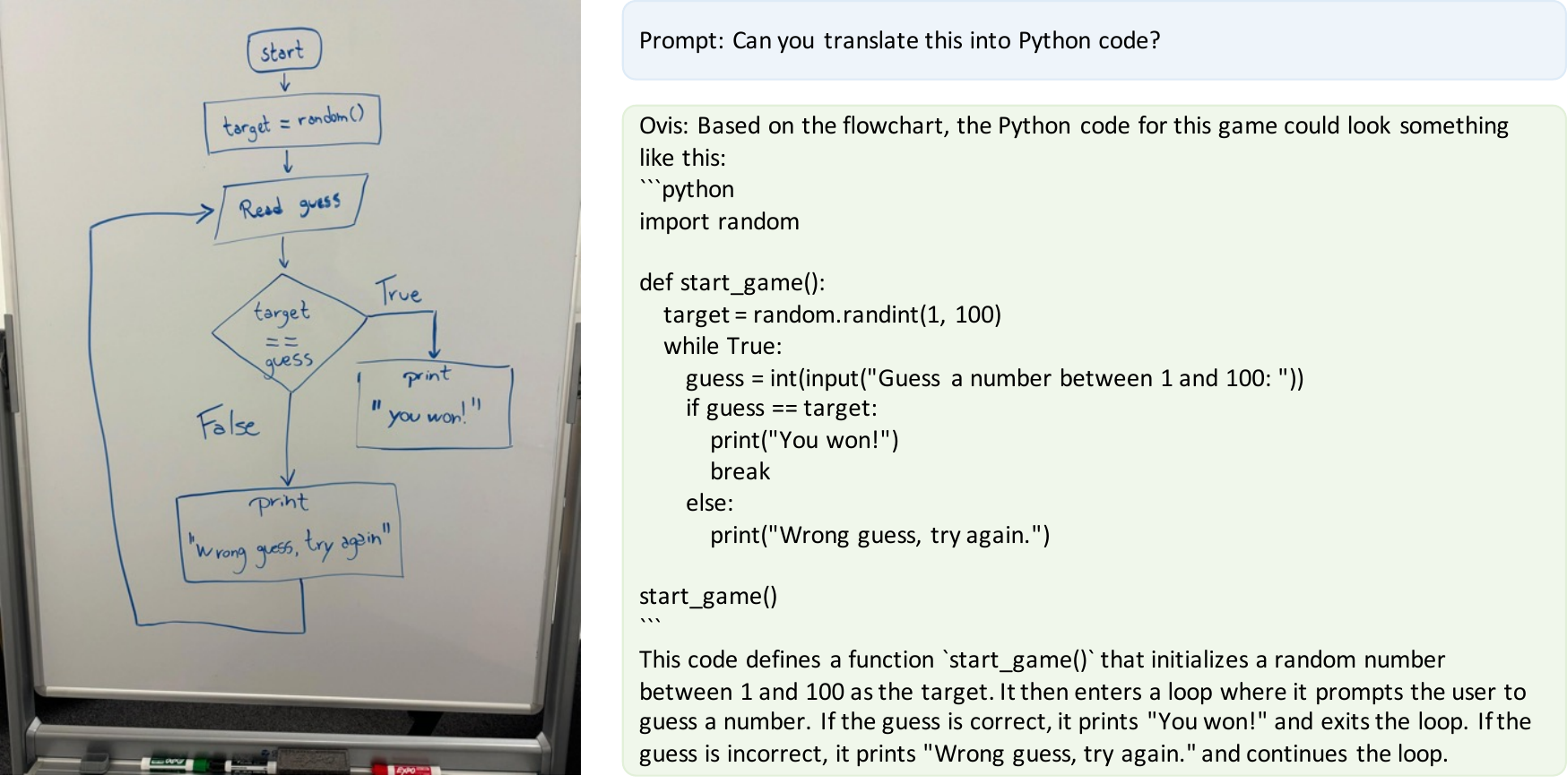}
    \caption{Coding \citep{grokv}}
    \label{fig:ovis-sample-coding}
  \end{subfigure}

  \caption{Qualitative results with Ovis-Llama3-8B, part I}
  \label{fig:ovis-samples}
\end{figure}

\begin{figure}[t]
  \centering
  \begin{subfigure}{\textwidth}
    \includegraphics[width=\linewidth]{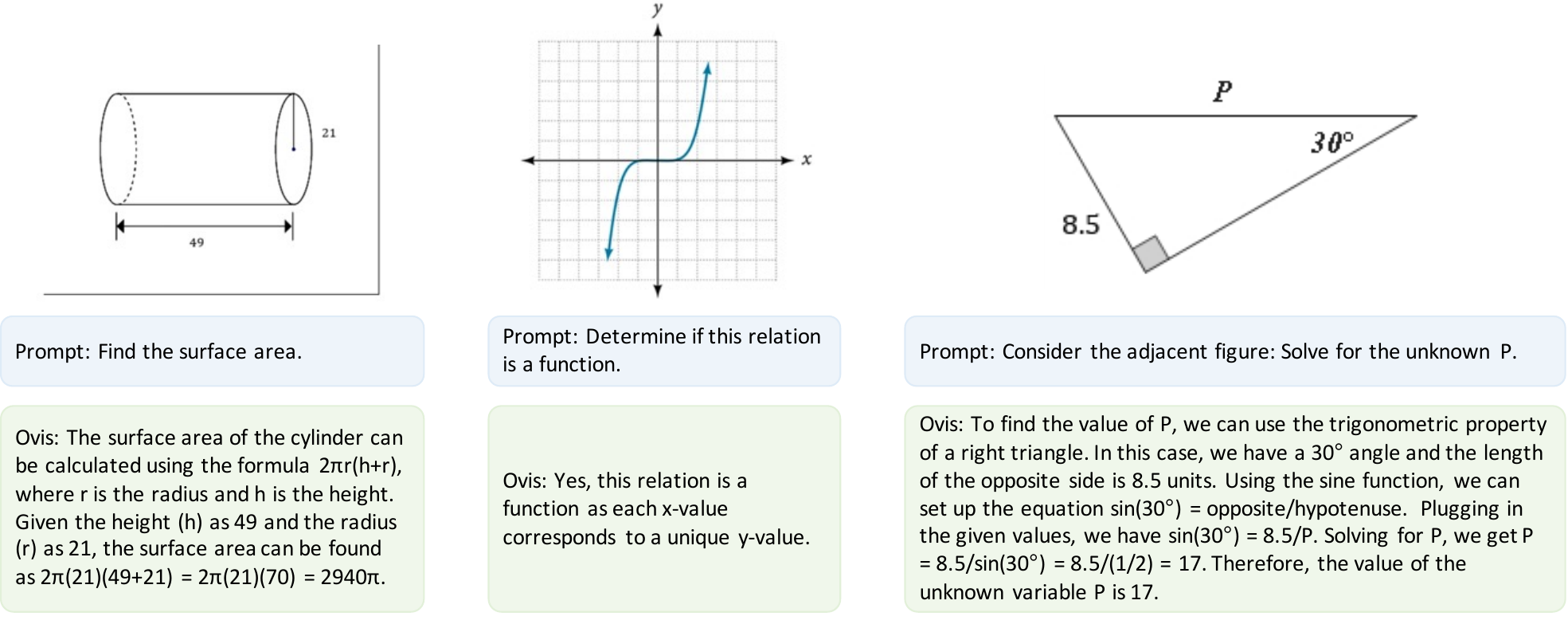}
    \caption{Math \citep{zhang2024mathverse}}
    \label{fig:ovis-sample-math}
  \end{subfigure}

  \hspace{1cm}
  
  \begin{subfigure}{\textwidth}
    \includegraphics[width=\linewidth]{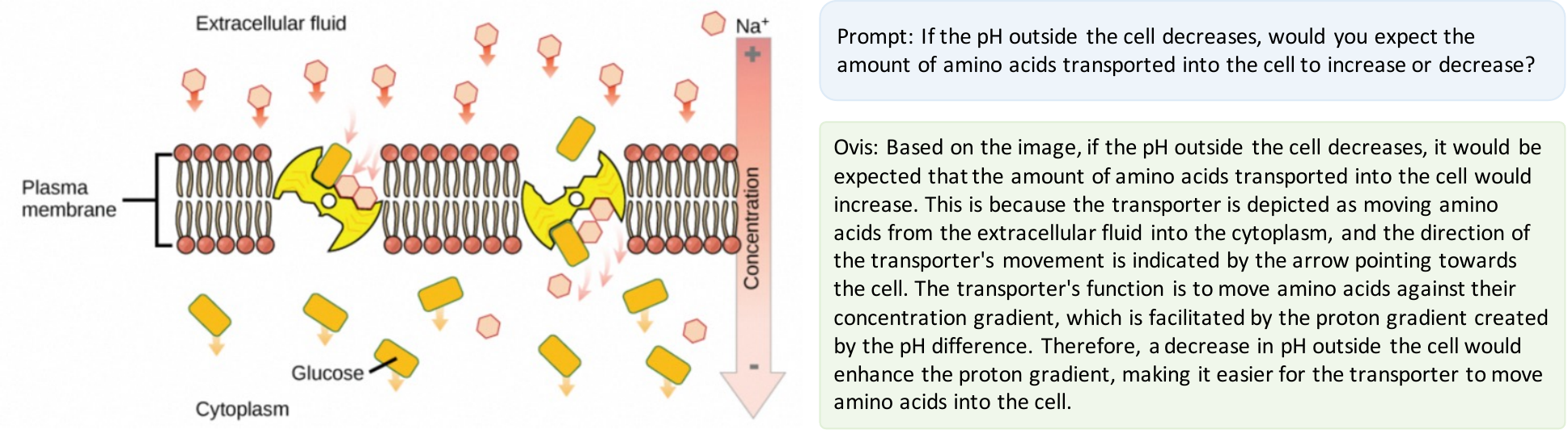}
    \caption{Science \citep{chen2024far}}
    \label{fig:ovis-sample-science}
  \end{subfigure}

  \hspace{1cm}
  
  \begin{subfigure}{\textwidth}
    \includegraphics[width=\linewidth]{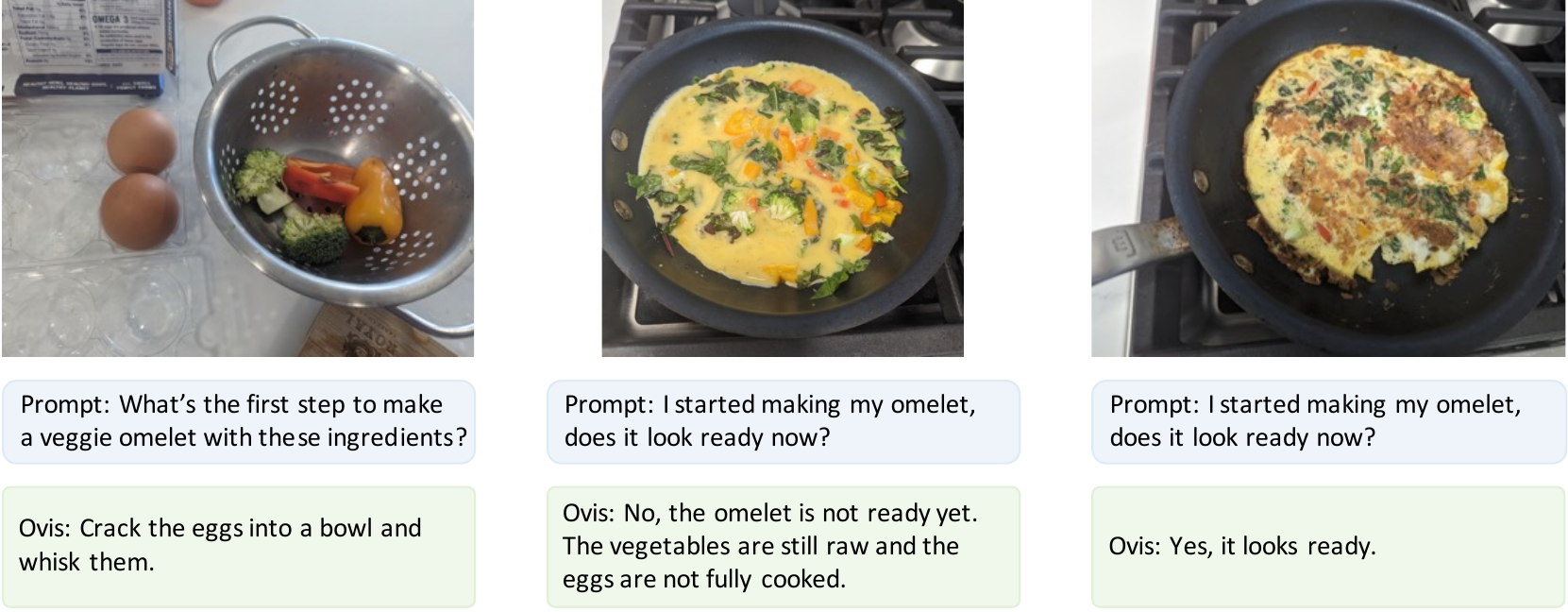}
    \caption{Cooking scenario \citep{Gemini}}
    \label{fig:ovis-sample-cooking}
  \end{subfigure}

  \caption{Qualitative results with Ovis-Llama3-8B, part II}
  \label{fig:ovis-samples-2}
\end{figure}

As shown in \autoref{fig:ovis-samples} and \autoref{fig:ovis-samples-2}, Ovis-Llama3-8B performs well in various multimodal tasks, where the images and prompts are sourced from literature.

\clearpage
\section{Training Details}
\label{sec:td}
The code and dataset for training are available at \url{https://github.com/AIDC-AI/Ovis}.
\begin{table}[h]
\centering
\caption{Training hyper-parameters}
\vspace{\baselineskip}
\label{tab:hp}
\begin{tabular}{@{}l|ccc@{}}
\toprule
Hyper-parameter  & Stage 1                        & Stage 2 & Stage 3 \\ \midrule
batch size      & 8192                           & 1024    & 1024                         \\
learning rate (Ovis-Qwen1.5-7B/14B)   & 1e-4                           & 1e-4    & 2e-5                         \\
learning rate (Ovis-Llama3-8B)   & 1e-4                           & 1e-4    & 1e-5                         \\
learning rate schedule        & cosine                         & cosine  & cosine                       \\
learning rate warm-up ratio   & 0.1                            & 0.1     & 0.05                         \\
weight decay    & 0                              & 0       & 0                            \\
grad norm clipping   & 1.0                            & 1.0     & 1.0                          \\
epoch           & 1                              & 1       & 1                            \\
optimizer       & AdamW                          & AdamW   & AdamW                        \\
float precision & bfloat16                           & bfloat16    & bfloat16                         \\
deepspeed configuration (Ovis-7B/8B)      & zero2 & zero3   & zero3                     \\
deepspeed configuration (Ovis-14B)      & zero3 & zero3   & zero3                     \\
training hours (Ovis-7B/8B, 128 H100 GPUs) & 6 & 2 & 7 \\
training hours (Ovis-14B, 128 H100 GPUs) & 10 & 6 & 21 \\ \bottomrule
\end{tabular}
\end{table}

\begin{table}[h]
\centering
\caption{Statistics of the training dataset with $\star$ denoting in-house data}
\vspace{\baselineskip}
\label{tab:data-stats}
\begin{tabular}{@{}l|lr|c@{}}
\toprule
Dataset Category                            & Dataset Name                                                           & \#Samples & Total Size         \\ \midrule
\multirow{1}{*}{Visual Caption}          & COYO-10M \citep{kakaobrain2022coyo-700m}              &  10M         &            10M        \\ \midrule
\multirow{6}{*}{Visual Description}      & LLaVA-Pretrain \citep{liu2023improved}                &       558K    & \multirow{6}{*}{2M}  \\
                                            & ShareGPT4V-Pretrain \citep{chen2023sharegpt4v}        &    82K       &                    \\
                                            & ALLaVA-Caption-Laion-4V \citep{chen2024allava}        &     485K      &                    \\
                                            & ALLaVA-Caption-Vflan-4V \citep{chen2024allava}        &     203K      &                    \\
                                            & Laion-Description$^\star$                                                      &   11K        &                    \\
                                            & CC12M-Description$^\star$                                                      &     1M      &                    \\ \midrule
\multirow{13}{*}{Multimodal Instruction} & ScienceQA-Train-Val \citep{lu2022learn}               &     17K      &  \multirow{13}{*}{3M}   \\ 
                                            & TextVQA-Train \citep{singh2019towards}                      &    35K                 &                    \\
                                            & ALLaVA-Instruct-Laion-4V \citep{chen2024allava}       &    485K               &                    \\
                                            & ALLaVA-Instruct-Vflan-4V \citep{chen2024allava}       &      203K     &                    \\
                                            & ArXivQA \citep{li2024multimodal}                      &    100K              &                    \\
                                            & Q-Instruct \citep{wu2023qinstruct}                    &    198K             &                    \\
                                            & LLaVA-Finetune \citep{liu2023improved}                &    665K             &                    \\
                                            & Geo \citep{gao2023g}                              &    177K              &                    \\
                                            & LRV-Instruction \citep{liu2023aligning}               &     300K               &                    \\
                                            & Chart-Instruction \citep{liu2023aligning}             &      43K              &                    \\
                                            & Synthdog-EN-OCR \citep{kim2022donut}                  &    200K              &                    \\
                                            & ALLaVA-Evol-Instruct \citep{chen2024allava} &    143K              &                    \\
                                            & CC12M-QA$^\star$                                                               &    387K                 &                    \\ \bottomrule
\end{tabular}
\end{table}

\clearpage
\section{In-house Visual Description Dataset}
\label{sec:id1}
\begin{figure}[htb]
	\begin{center}
		\includegraphics[width=1\columnwidth]{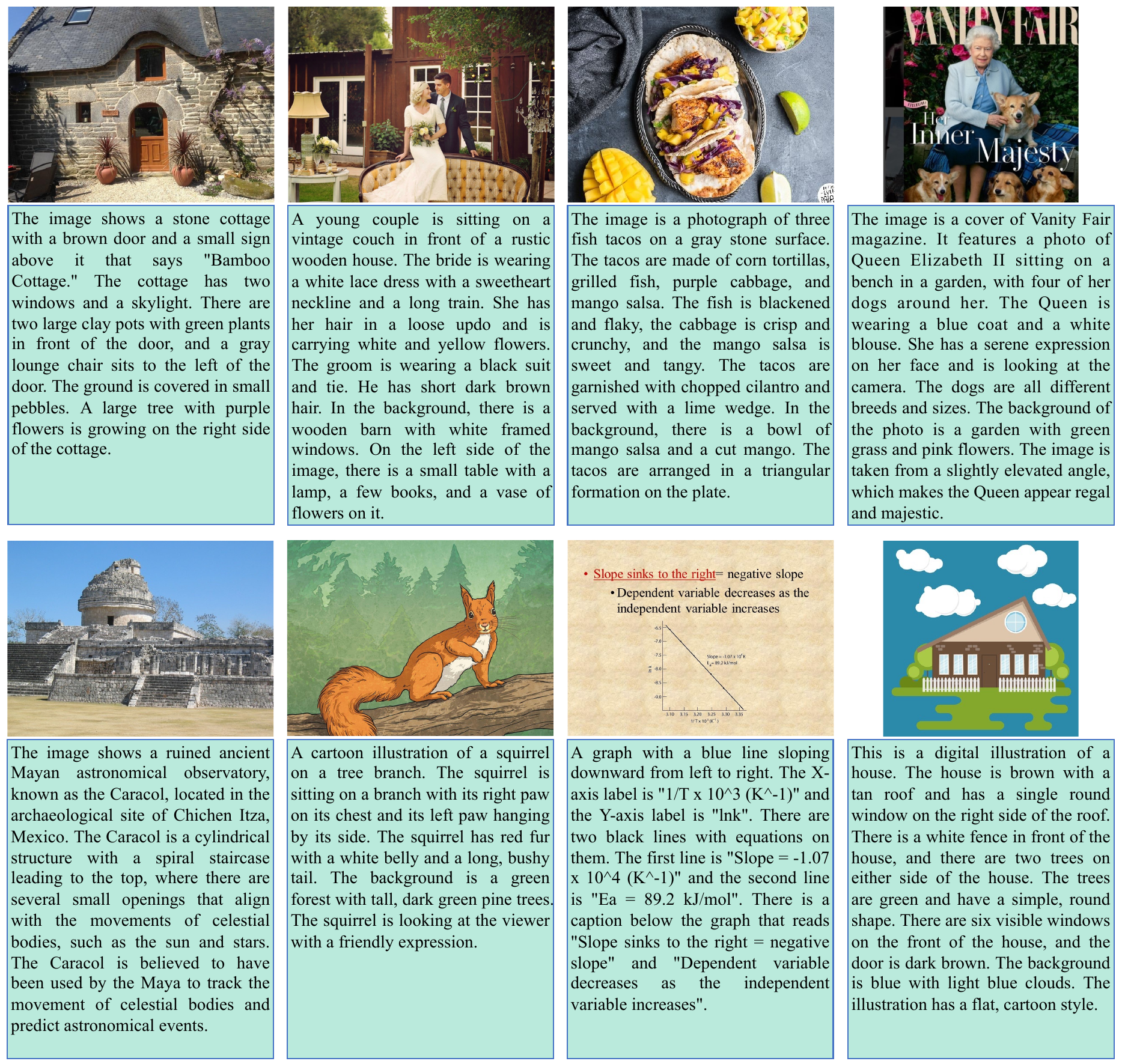}
	\end{center}
	\caption{Samples from our in-house visual description dataset} \label{figure:InHouseDataS2}
\end{figure}

We sample images from the Laion \citep{laion-5b} and CC12M \citep{changpinyo2021cc12m} datasets, which cover various categories, including nature, lifestyle, humanities, architecture, cartoons, and abstract art, as shown in Figure~\ref{figure:InHouseDataS2}. For each image, we call the Gemini-Pro or GPT-4V API with a unified prompt to generate the image's descriptions. The unified prompt explicitly requires the API to reply with concise and clear visual information about the image, as well as performing OCR recognition if relevant, while avoiding embellishments and interpretations.

\clearpage
\section{In-house Visual Instruction Dataset} 
\label{sec:id2}
\begin{figure*}[htb]
	\begin{center}
		\includegraphics[width=1\columnwidth]{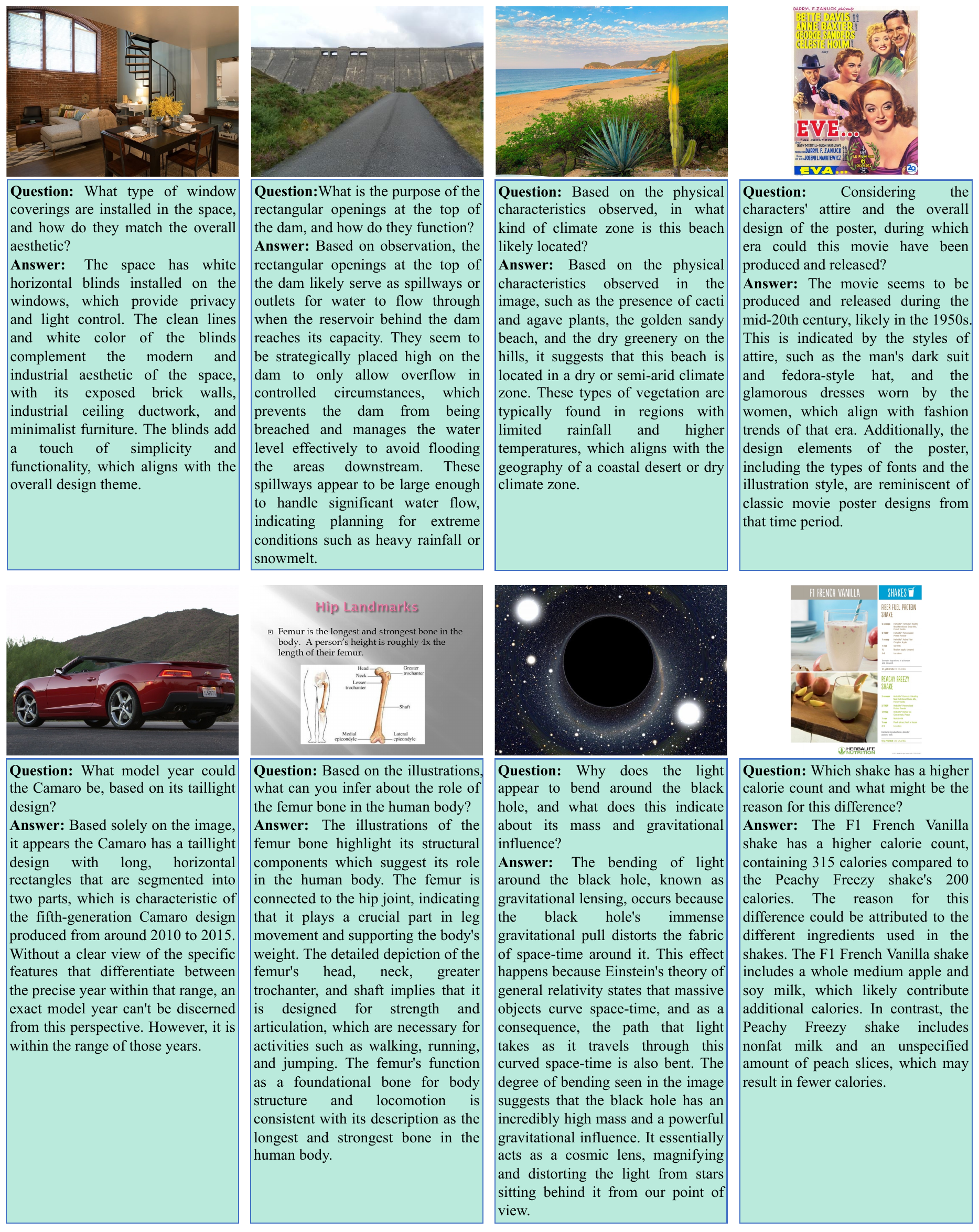}
	\end{center}
	\caption{Samples from our in-house visual instruction dataset} \label{figure:InHouseDataS3}
\end{figure*}

We create visual instruction samples for images from the CC12M dataset \citep{changpinyo2021cc12m} in a similar way to \citep{chen2024allava}, using Gemini-Pro and GPT-4V to carry out a self-questioning and answering task. This method produces diverse questions and high-quality answers, as illustrated in Figure~\ref{figure:InHouseDataS3}.

\clearpage
\section{Sparsity of Ovis' Visual Tokenizer}\label{sec:appendix-sparsity}
\begin{figure*}[htb]
	\begin{center}
		\includegraphics[width=0.6\columnwidth]{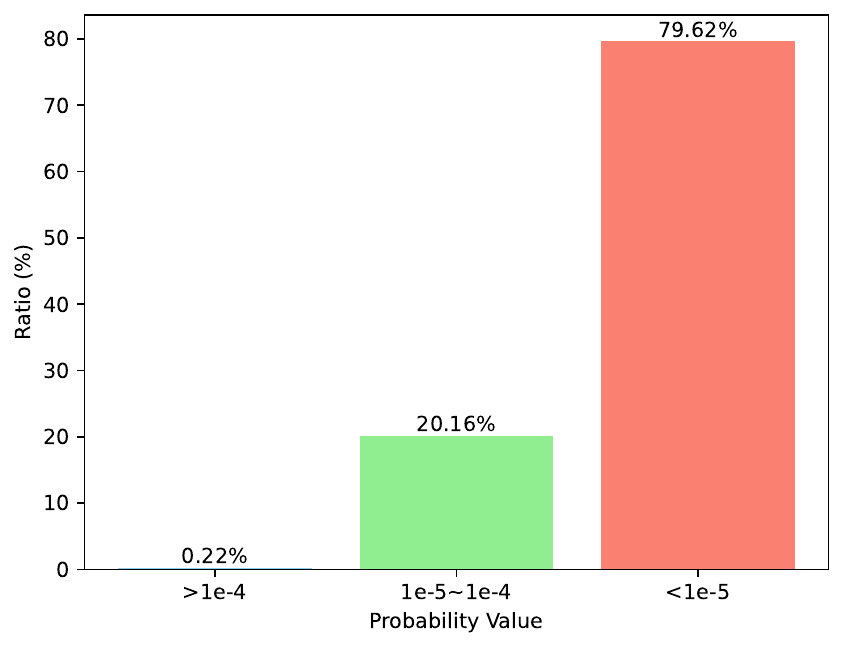}
	\end{center}
	\caption{Statistics of the probability distributions characterized by the visual tokens across 10,000 images from ImageNet-1K. The visual tokens are obtained using Ovis-Llama3-8B's visual tokenizer.} \label{figure:vt-sparsity}
\end{figure*}

To assess the sparsity of Ovis' visual tokenizer, we conduct an experiment using 10,000 images sampled from the ImageNet-1K dataset \citep{imagenet15russakovsky}. Each image is tokenized by the visual tokenizer of Ovis-Llama3-8B, resulting in a sequence of visual tokens, each of which is a probability distribution over the visual vocabulary. We then employ thresholds of 1e-4, 1e-5, and 1e-6 to categorize the probability values and calculate the ratio of values falling within each interval across the 10,000 images. As shown in \autoref{figure:vt-sparsity}, the probability distributions characterized by the visual tokens are highly sparse, with only 0.22\% probability values exceeding the threshold of 1e-4.

\end{document}